\documentclass[11pt]{article}

\usepackage[final]{acl}

\usepackage{times}
\usepackage{latexsym}

\usepackage[T1]{fontenc}

\usepackage[utf8]{inputenc}

\usepackage{devanagari}
\usepackage{enumitem}
\usepackage{microtype}
\usepackage{inconsolata}
\usepackage{etoolbox}
\AtBeginDocument{\setbox0=\hbox{\texttt{}}}
\usepackage{float}
\usepackage{graphicx}
\usepackage{amsmath}
\usepackage{cleveref}
\usepackage{amsmath,amsfonts,bm}

\def\eqref#1{equation~\ref{#1}}
\def\1{\bm{1}}

\DeclareMathAlphabet{\mathsfit}{\encodingdefault}{\sfdefault}{m}{sl}
\SetMathAlphabet{\mathsfit}{bold}{\encodingdefault}{\sfdefault}{bx}{n}

\DeclareMathOperator*{\argmax}{arg\,max}

\title{Target-Language Generation in Multilingual Models:\\Activation Steering and Optimal Control }

\author{
 James A. Michaelov
  \\
  Massachusetts Institute of Technology
  \And
  Carmen Amo Alonso 
  \\
  Stanford University
  \AND
 Tyler A. Chang
  \\
  University of California San Diego
  \And
  Roger P. Levy 
  \\
  Massachusetts Institute of Technology
}

\date{}

\begin{document}
\maketitle

\begin{abstract}
    Ensuring that multilingual language models generate coherent text in a specific target language is a major issue in multilingual language modeling. We develop an optimal control method for target-language text generation as well as a framework for evaluating the quality of generated text in terms of language adherence, linguistic coherence, and semantic coherence. We find that the proposed method performs at least as well as the prominent difference-in-means activation steering method for the majority of models tested, with substantially less hyperparameter tuning required. 

   \raisebox{-2pt}{\includegraphics[scale=0.09]{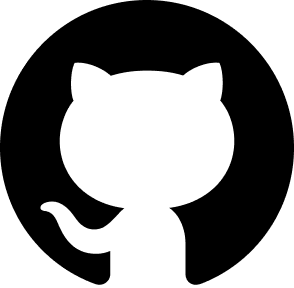}}  \href{https://github.com/jmichaelov/target-language-control}{jmichaelov/target-language-control}

\end{abstract}

\section{Introduction}
\label{sec:intro}

\begin{figure*}[h]
    \centering
    \includegraphics[width=\linewidth]{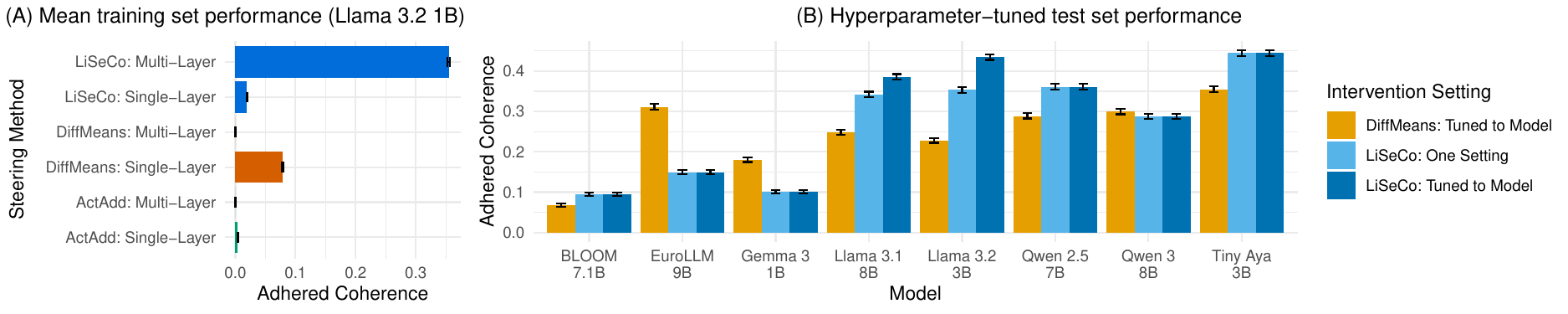}
    \caption{(A) Performance of Llama 3.2 1B after activation steering (on all English, Spanish, and Hindi language pairs of the \textsc{XStoryCloze} training set), averaged across all \textsc{ActAdd}, \textsc{DiffMeans}, and \textsc{LiSeCo} hyperparamters. (B) Performance of 8 language models after activation steering (on all English, Spanish, Hindi, and Chinese language pairs of the \textsc{XStoryCloze} test set), using the single-layer \textsc{DiffMeans} and multi-layer \textsc{LiSeCo}, with hyperparameters tuned to a single setting (\textsc{LiSeCo}), or to each individual language model (\textsc{DiffMeans}, \textsc{LiSeCo}). Error bars reflect 95\% bootstrapped confidence intervals.}
    \label{fig:summary_fig}
\end{figure*}

Despite substantial advances in natural language technology in recent years, the vast majority of progress has focused on a limited number of high-resource languages, primarily English \citep{joshi-etal-2020-state,nigatu-etal-2024-zenos,qin_survey_2025,wu2025bitterlessonlearned2000}. While capabilities can transfer across languages, current research suggests that the extent to which this is the case is limited and uneven across languages and capabilities \citep{conneau_unsupervised_2020,wang_negative_2020,chang_when_2023}. One known issue is that when prompted in a specific target language or explicitly prompted to generate text in that language, even state-of-the-art models can drift into generating in higher-resource languages, especially when generating longer texts or during reasoning \citep{marchisio_understanding_2024,yong_2025_CrosslingualReasoningTestTime}.

In this paper, we address the multifaceted challenge of generating text in a specific target language. First, we consider the difficulty of evaluating whether a language model has successfully done so. We identify three key desiderata for target-language generation, namely, \textbf{language adherence} (whether the model generates text in the target language), \textbf{linguistic coherence} (whether the model generates a valid string of the target language), and \textbf{semantic coherence} (whether the model generates text with the appropriate meaning and relation to its context, e.g., the prompt). We propose a specific operationalization of each of these desiderata as metrics, and develop a metric that combines them appropriately, which we name \textbf{adhered coherence}.

Next, we narrow our focus to one specific and promising avenue for target-language generation, namely, activation steering. We formulate this approach to target-language generation as an optimal control problem based on the framework developed by \citet{cheng2024linearly}, and use this to formally compare the accompanying Linear Semantic Control method of activation steering (\textsc{LiSeCo}; \citealp{cheng2024linearly,cheng2026liseco}) to approaches that have been proposed for target-language activation steering, including both the standard difference-in-means method (\textsc{DiffMeans}; \citealp{rimsky_steering_2024}), and other more specific approaches tailored to language steering. Based on this, we argue that a \textsc{LiSeCo}-based approach has desirable properties in that it is in principle more flexible and provides formal guarantees that other methods do not.

We then empirically compare methods. We adapt the \textsc{LiSeCo} method of \citet{cheng2024linearly} for target-language activation steering, and compare this to \textsc{DiffMeans}, as well as the related lightweight \textsc{ActAdd} method. Using crosslingually-paired \textsc{XStoryCloze} items, we then evaluate the performance of each method at switching the language of text generation without any prompt (i.e., `language forcing'; \citealp{gurgurov2025language}). While we observe substantial variation by model and language pair, we find that \textsc{LiSeCo} empirically performs more predictably and generally better than the \textsc{DiffMeans} and \textsc{ActAdd} approaches, and requires substantially less hyperparameter tuning to be effective.

\section{Related Work}
\label{sec:related_work}

\subsection{Multilingual Representations and Target-Language Activation Steering}

While there is some evidence that some semantic knowledge is best accessed in the language in which it is learned \citep{mittal_2023_MOKB6MultilingualOpen,ifergan_2025_SurfaceConsistencyExploring,goldman_eclektic_2025,zhong_2025_WhatLanguageNonEnglishCentric}, an increasing amount of evidence suggests that language representations and semantic representations can often be dissociated, particularly for languages on which models have been more extensively trained \citep{tezuka_2025_TransferNeuronsHypothesis,chen_2025_EmergenceAbstractThought,tamo_2025_LinguaMapWhichLayers}. For example, activating neurons associated with a concept in one language can lead to models generating text related to that concept in another language \citep{riemenschneider2025cross}, and activating neurons associated with a specific language can be used to generate text in that language \citep{gurgurov2025language,sundar2025steering}. Similarly, the difference-in-means (\textsc{DiffMeans}; \citealp{subramani2022extracting,li2023inferencetime,rimsky_steering_2024,marks_geometry_2024}) method, where the difference between mean representations of texts with two different features is used to steer the model in the direction of either has been applied to language \citep{lu_paths_2025}, as have other approaches that involve calculating or learning a mapping between two languages \citep{chang_2022_GeometryMultilingualLanguage,mahmoud2025improving,wang2025bridging,sterz_2025_ReCoVeRTargetLanguage,lopo_2025_LanguageSurgeryMultilingual}.

Because Transformer model representations are highly anisotropic \citep{ethayarajh_2019_HowContextualArea,hammerl_2023_ExploringAnisotropyOutliers}, one possible issue with activation steering is that the model's activation state could be shifted into an `unnatural' subspace, where it is within the right region along the dimensions for desired language, but other dimensions of the representations could be degraded. Thus, most of the aforementioned methods try to reduce the extent to which representations are shifted during steering, for example by reducing the distance in activation space to which representations are steered \citep[e.g.,][]{wang2025bridging}, only intervening on a small subset of layers \citep[e.g.,][]{lu_paths_2025}, only intervening on a small subset of neurons \citep[e.g.,][]{gurgurov2025language}, or other similar approaches. Generally, the research suggests that input-language-specific representations are most distinct in the earlier layers, and output-language-specific representations in the later layers, with more language-neutral representations in the middle layers \citep{kojima2024multilingual,wang2025lost,dumas2024separating,zhang2025same,wu_semantic_2025,tezuka_2025_TransferNeuronsHypothesis}. Thus, when the goal is to steer the model to generate text in a specific language, one would expect that intervening on later layers is likely to be the most successful approach \citep[see][]{lu_paths_2025}.

\subsection{Evaluation}
Previous work evaluating target-language generation through activation steering generally focuses on one or more of three settings. The first is prompting a model in a target language and evaluating whether the activation intervention leads to an improvement in text generated in the same language, either due to intervening to align the model's internal representations more with a high-resource language such as English \citep{wang2025bridging,lu_paths_2025,mahmoud2025improving} or intervening to steer the model's output in the direction of the target language \citep{sterz_2025_ReCoVeRTargetLanguage}. The second is prompting a model to generate text in a target language and evaluating whether it successfully generates text in that language \citep{sterz_2025_ReCoVeRTargetLanguage,lopo_2025_LanguageSurgeryMultilingual}. The third common type of experiment is to intervene in some way on the model's representations to attempt to cause it to generate text in a target language and evaluating whether it does so successfully \citep{chang_2022_GeometryMultilingualLanguage,sundar2025steering,gurgurov2025language,gurgurov-etal-2026-clas}.

One of the primary evaluation metrics in such experiments is whether the language model generates text in the appropriate language, often referred to as \textit{language adherence} \citep{langlais2024they}. In general, the approach taken is to use a language identification classifier on the generated text, and a model is considered to adhere if the generated text is assigned the label corresponding to the target language \citep{chang_2022_GeometryMultilingualLanguage,sundar2025steering,gurgurov2025language}. This can be calculated at the level of a whole text generation, or in a more fine-grained fashion by classifying individual words or lines of the text to see what proportion successfully adhere (see \citealp{marchisio_understanding_2024}).

Alternatively or in addition, generations can be evaluated based on the quality of the input and output. For example, activation steering has been used to improve translation quality \citep{lopo_2025_LanguageSurgeryMultilingual,wang2025bridging}; in this case, evaluation based on a reference translation inherently captures language adherence. Cases where the models are evaluated on tasks where there is a reference or ground-truth answer in the target language (as in the case of same-language prompting or including the generation language in the prompt) can also be treated in the same way \citep{mahmoud2025improving,lu_paths_2025,wang2025bridging}, though some work evaluates language adherence separately in addition to task accuracy \citep{sterz_2025_ReCoVeRTargetLanguage,lopo_2025_LanguageSurgeryMultilingual}. Finally, in contemporaneous work, \citet{gurgurov-etal-2026-clas} fully separate language adherence and content (in terms of both linguistic and semantic coherence as defined in \Cref{sec:evaluation}) by evaluating generation quality using a multilingual LLM-as-judge that is instructed to ignore language adherence in its rating.

\section{Evaluating language activation steering}
\label{sec:evaluation}
Evaluating target-language text generations is a multifaceted problem that requires both assessing whether the generated text is in the correct language, and assessing the appropriateness of the generated text itself. We develop a principled approach to assess this. Following previous work on summary evaluation \citep{peyrard2017learning,zhu2020gruen}, we begin by determining three key desiderata that we believe are important to evaluating model generations in context.
We then describe how they can be combined into a single principled metric of target-language text generation quality.

\paragraph{Language Adherence} When assessing the quality of the text generated by a model in a specific target language, perhaps the most straightforward aspect to test is whether the text is in fact in the target language. We consider language adherence as a binary variable, specifically, whether a language classifier that predicts the language $\mathcal{L}_i$ of a given text assigns target language $t$ the highest probability given the generated text (i.e., $\argmax_i\hat{p}(\mathcal{L}_i) =t$).

\paragraph{Semantic Coherence} In addition to language adherence, it is also important for the generated text to have appropriate content given its context. For our purposes, we consider a specific case where the output is not straightforwardly verifiable, and the task has reference correct ($s_{\text{true}}$) and incorrect ($s_{\text{false}}$) outputs. Thus, to get a binary value for semantic coherence for a given output string $s_{\text{gen}}$, we can calculate whether $\text{sim}(s_{\text{gen}},s_{\text{true}})>\text{sim}(s_{\text{gen}},s_{\text{false}})$, where similarity is calculated from multilingual sentence embeddings.

\paragraph{Linguistic Coherence} Less considered in previous work is the question of whether the generated text is not only recognizable as being in a given language and contains semantically appropriate content, but is also well-formed in the target language. We operationalize \textit{linguistic coherence} as log-perplexity. We specifically propose the use of $n$-gram perplexity because it avoids the confounds of evaluating a model's output based on another model with the same architecture---for example, a common failure mode for neural language models is repetition \citep{fu_2021_TheoreticalAnalysisRepetition,xu_2022_LearningBreakLoopa,li_2023_RepetitionRepetitionOut,hiraoka_2025_RepetitionNeuronsHow,doan_2025_UnderstandingControllingRepetition}, and if a model susceptible to this were used to evaluate the generations, it would lead to repetitive generations being assigned unduly low (i.e., good) perplexities. For our purposes, as with the other two metrics, we treat linguistic coherence as a binary variable by setting a threshold value. We set a linguistic coherence threshold such that a response's log-perplexity in the target language $L_t$ must have a value smaller than than 2 standard deviations above the mean baseline log-perplexity in the target language $L_b$, where the baseline refers to a monolingual setting in the target language with no activation steering. Thus, our measure of linguistic coherence is described by the equation $L_t\leq\mu(L_b)+2\sigma(L_{b})$.

\paragraph{Adhered coherence} We then combine these three metrics in a joint score that encompasses all three desiderata, which we name \textit{adhered coherence}. As shown in \Cref{eq:adhered_coherence}, a specific text generation demonstrates adhered coherence $C$ if it adheres to the target language, is semantically coherent, and is linguistically coherent, as defined previously in this section:

\begin{equation}\label{eq:adhered_coherence}
    C = \begin{cases} 
      1 & \text{if } \argmax_i\hat{p}(\mathcal{L}_i) =t\\ 
      & \text{and}~ \text{sim}(s_{\text{gen}},s_{\text{true}})>\text{sim}(s_{\text{gen}},s_{\text{false}})\\
      & \text{and}~L_t\leq\mu(L_b)+2\sigma(L_{b})\\ \vspace{-0.25cm}\\
      
      0 & \text{otherwise} 
   \end{cases}
\end{equation}

\section{Target-Language Generation as an Optimal Control Problem}
\label{sec:problem_formulation}

\subsection{Activations Across Layers as Dynamic Trajectories}

Following \citet{cheng2024linearly}, one can conceptualize the flow of activations across Transformer layers as forming dynamic trajectories through high-dimensional representation spaces. Transformations of the input representations are realized in a sequential manner given the iterative nature of language model layers. Hence, if we treat the activation state at a given token as capturing the representation of the current token and relevant information from previous tokens (as in, e.g., \citealp{olsson_2022_IncontextLearningInduction,rimsky_steering_2024,wang2025bridging}), the trajectory of an activation through layers can be described as a discrete-time dynamical system propagation of the form:
\begin{multline}
\label{eqn:dynamics}
    x_0 =  E(s_i), \quad x_{\tau+1} = \ell_{\tau+1}(x_\tau), \\s_{i+1} = U(x_T), \\ \text{with } \tau=0,\dots,T-1 
\end{multline}
where $s\in\Sigma^*$ is the prompt string, $x_\tau\in\mathbb R^d$ is the latent representation of string $s$ after layer $\ell_\tau$, $\ell_\tau$ is the $\tau^{th}$ model layer, $T$ is the number of layers in the model, and $E$ and $U$ are the \textit{embedding} and \textit{unembedding} matrices, respectively.

Evidence suggests that cross-lingual understanding emerges from the geometry and dynamics of these activation spaces naturally, rather than from explicit training objectives for cross-lingual alignment (see \S\ref{sec:related_work}). Thus, the dynamical systems perspective above offers a principled framework for understanding and manipulating multilingual generation: by viewing the Transformer as implementing the dynamics in Equation~\ref{eqn:dynamics}, we can conceptualize cross-lingual steering as a trajectory control problem, where interventions guide the activation flow toward desired linguistic or semantic targets while preserving the underlying conceptual structure that enables cross-lingual generalization.

\subsection{Problem Statement}

Building on the dynamical systems perspective, we formulate cross-lingual steering as an intervention design problem in activation space. Given a trained multilingual language model and a source language input, the goal is to design activation interventions that guide the model's output toward a target language while preserving semantic content and maintaining generation quality. Given a target language $\mathcal{L}_t$, let $\mathcal{R}_t \subset \mathbb{R}^d$ represent the region in layer $t$'s activation space corresponding to generations in that target language. The goal is to design control inputs $\theta_\tau: \mathbb{R}^d \to \mathbb{R}^d$ to modify the dynamics of the last token's activations:
\begin{multline}
    x_0 = E(s), \quad \tilde{x}_\tau = x_\tau + \theta_\tau(x_\tau), \\ x_{\tau+1} = \ell_{\tau+1}(\tilde{x}_\tau), \quad y = U(x_T),
\end{multline}
such that the following requirements are satisfied:

\begin{itemize}[itemsep=0cm]
\item \textbf{Minimal Disturbance}: The intervention should introduce the smallest possible perturbation to the original activation trajectory, i.e., should prevent oversteering and preserve the model's learned representations.
    
    \item \textbf{Guaranteed Language Transition}: The modified activations must reliably steer the language attribute toward the target, ensuring that $\tilde{x}_\tau \in \mathcal{R}_t$ for intervened layers $\tau \in \mathcal{T}$, such that for any text in source language $\mathcal L_s$, the generated continuation is classified to be in target language $\mathcal{L}_t$, where $\mathcal{T}$ is the set of layers chosen for intervention.
    
    \item \textbf{Topology Awareness}: The intervention must respect the geometric structure of the embedding space, leveraging the multilingual topology to move between language-specific regions while maintaining semantic coherence and conceptual alignment with the original source language.
\end{itemize}

The challenge lies in simultaneously satisfying these competing objectives: achieving reliable language transition while minimizing activation perturbations and respecting the learned multilingual geometry underlying the model's core capabilities.

\subsection{Existing Approaches} 
\label{ssec:existing_approaches}
Our formulation allows us to describe current methods in terms of $\theta_\tau$. With \textsc{DiffMeans} (see, e.g., \citealp{lu_paths_2025}), $\theta_\tau=\kappa z_t$, where $z_\tau=\mu_\tau^{\mathcal{L}_t}-\mu_\tau^{\mathcal{L}_s}$ is computed from paired data in the source and target languages and $\kappa$ is a hyperparameter. The neuron manipulation approach \citep{gurgurov2025language} also has $\theta_\tau=\kappa z_\tau$, but in this case $z_\tau$ is a sparse vector ($\leq$5\% of neurons are intervened on) where non-zero values are `boost values' calculated for each individual neuron based on $\mu_\tau^{\mathcal{L}_t}$. Most other previous studies (e.g., \citealp{chang_2022_GeometryMultilingualLanguage,sundar2025steering,mahmoud2025improving,wang2025bridging,sterz_2025_ReCoVeRTargetLanguage,lopo_2025_LanguageSurgeryMultilingual}) can generally be considered variants of one (or a combination) of these approaches---crucially, under these approaches $\theta_\tau$ does not depend on the original activation $x_\tau$. The exception to this is INCLINE \citep{wang2025bridging}, where $\theta_\tau=\kappa x_\tau V_\tau$, and $V_\tau$ is a learned matrix that maps $x_\tau^{\mathcal{L}_s}$ to $x_\tau^{\mathcal L_t}$, fitted based on paired data in the source and target languages.

\section{Target-Language Interventions in Activation Space via Optimal Control}
\label{sec:multilingual_control}

\begin{table*}[h!]
    \centering
    \renewcommand{\arraystretch}{2}
    \setlength{\tabcolsep}{15pt}
    \begin{tabular}{|c||c|c|c|}
        \hline
        \textbf{Condition} & $\sigma(W_\tau^T x_\tau) > \alpha^{\text{max}}$ & $\sigma(W_\tau^T x_\tau) < \alpha^{\text{min}}$ & \text{otherwise} \\ 
        \hline\hline
        $\boldsymbol{\theta_\tau^*}$ & $\displaystyle\frac{\log(\frac{1}{\alpha^{\text{max}}}-1) - W_\tau^T x_\tau}{\|W_\tau\|_2^2}W_\tau$ & $\displaystyle\frac{\log(\frac{1}{\alpha^{\text{min}}}-1) - W_\tau^T x_\tau}{\|W_\tau\|_2^2}W_\tau$ & $0$ \\
        \hline
    \end{tabular}
    \caption{Optimal intervention $\theta_\tau^*$ for language control at layer $t$.}
    \label{table:theta_t_star}
\end{table*}

\subsection{Multilingual Region Identification in Activation Space}

Identifying the language-specific region $\mathcal{R}_t^i$ depends on how the language attribute $a$ is encoded in latent space. Let layer $t$'s activations encode language as $f_t: \mathbb{R}^d \to \mathcal{L}$, mapping activation vectors to discrete language labels. Then, the region $\mathcal{R}_t^i$ in latent space can be identified as the pre-image of language $\mathcal{L}_i$ under $f_t$. This information can then be leveraged to control activations, since the desired language outcome $a = \mathcal{L}_t^*$ can be proxied by enforcing the activation $x_t \in \mathcal{R}_t^{\mathcal{L}_t^*}$.

\paragraph{Linear Representation Hypothesis. \cite{park2023the}} In the Linear Separability Hypothesis, each language $\mathcal{L}_i$ is hypothesized to occupy a distinct region $\mathcal{R}_t^i$ of activation space for each layer $t$. Formally, a string $s\in\Sigma^*$ generates text in language $\mathcal{L}_i$ if and only if its corresponding representations $x_t\in\mathcal{R}_t^i$, where $\mathcal{R}_t^i$ is a linearly-separable region in embedding space. 

The goal is to identify the regions in activation space that correspond to different languages. Specifically, at each layer $t$ we learn a lightweight linear probe $f_t$ that maps the latent state $x_t$ to language probabilities. By the linearity hypothesis, we define $f_t: \mathbb{R}^d \to \mathbb{R}^{|\mathcal{L}|}$ as:
\begin{equation}\label{eqn:language_probe}
    f_\tau(x_t) = \text{softmax}(W_\tau^T x_\tau),
\end{equation}
where $W_\tau \in \mathbb{R}^{d \times |\mathcal{L}|}$ is a weight matrix that projects activations to language logits. 

For each layer $\tau$, we minimize the cross-entropy loss over a dataset $\{s^{(i)}, \mathcal{L}^{(i)}\}_{i=1}^N$ of (text, language) pairs:
\begin{equation}\label{eqn:language_loss}
    \min_{W_t} -\sum_{i=1}^N \sum_{j=1}^{|\mathcal{L}|} \mathbf{1}[\mathcal{L}^{(i)} = \mathcal{L}_j] \log f_\tau(x_\tau^{(i)})_j
\end{equation}
where $x_\tau^{(i)}$ is the representation of string $s^{(i)}$ at layer $\tau$, and $f_\tau(x_\tau^{(i)})_j$ is the predicted probability for language $\mathcal{L}_j$. 

\subsection{Multilingual Control of Activations}

Having identified language-specific regions in activation space, we now turn to the problem of controlling activations to steer generation toward the target language.

\paragraph{Semantic Hub Hypothesis \cite{wu_semantic_2025}.} The Semantic Hub Hypothesis provides crucial insight into how multilingual control should be designed. It posits that semantically equivalent inputs $s^{\mathcal{L}_s}$ and $s^{\mathcal{L}_t}$ from source and target languages have representations such that:
\begin{equation}
    \text{sim}(x^{\mathcal{L}_s}, x^{\mathcal{L}_t}) > \text{sim}(x^{\mathcal{L}_s}, u^{\mathcal{L}_t}),
\end{equation}
where $\text{sim}$ is a similarity measure and $u^{\mathcal{L}_t}$ represents semantically unrelated content in the target language.
In other words, semantically equivalent content in the target language lies closer to the source representation than does semantically unrelated content in the target language; thus, the \textit{nearest} point in the target language region is a representation that preserves semantic content.
This suggests that in order for cross-lingual control to preserve semantic content while transitioning between language-specific regions, it is enough to compute \emph{minimal} interventions to $x^{\mathcal{L}_s}$ as long as the modified activation is guaranteed to lie in $\mathcal R^t$.

The goal is to design control interventions $\theta_\tau$ that respect this geometric structure. Therefore, we formulate the control problem as finding the minimal intervention that moves activations into the target region $\mathcal{R}_t^{\mathcal{L}_t^*}$. Following the framework of \textsc{LiSeCo} \cite{cheng2024linearly}, we can adapt their continuous attribute control formulation to our discrete language setting. For binary language classification (source vs. target), the control problem becomes:
\begin{subequations}\label{eqn:relaxed_control_continuous_range}
    \begin{align}
    \underset{\theta_\tau}{\text{min}} & \qquad \|\theta_\tau\|_2^2 \label{eqn:global_cost_relx_cont_range}\\
    s.t. 
    &  \qquad \alpha^{\text{min}} \leq f_\tau(x_\tau + \theta_\tau)) \leq \alpha^{\text{max}},
    \label{eqn:constr_probe_relx_range} 
    \end{align}
\end{subequations}
for each layer $\tau \in \mathcal{T}$ and $[\alpha^{\text{min}}, \alpha^{\text{max}}]$ defines the desired probability range for the target language.

A key advantage of this formulation is that it admits a closed-form solution, enabling efficient computation with minimal overhead. From Theorem $1$ in \citet{cheng2024linearly}, the optimal solution $\theta_\tau^*\in\mathbb{R}^d$ to the optimization problem \ref{eqn:relaxed_control_continuous_range} is given in Table \ref{table:theta_t_star}.
Geometrically, the optimal solution projects $x_t$ onto the closest point in the target language region $\mathcal{R}_t^{\mathcal{L}_t^*}$. When the current activation already lies within the desired region, no intervention is needed ($\theta_\tau^* = 0$). Otherwise, the intervention is a scaled version of the probe direction $W_\tau$, with magnitude determined by the distance to the region boundary. This approach guarantees that the perturbed activation will lie within the target language region while minimizing the perturbation magnitude.

The proposed intervention, based on optimal control, provides a principled framework that is more general and flexible than existing multilingual intervention methods. For example, as discussed in \S\ref{ssec:existing_approaches}, under most existing methods $\theta_\tau$ does not depend on $x_\tau$, but rather is fixed. Furthermore, the one existing method where it does (INCLINE; \citealp{wang2025bridging}) relies on mapping from $x_\tau^{\mathcal L_s}$ to $x_\tau^{\mathcal L_t}$ based on a pre-trained alignment matrix $V_\tau$ and shifting $x_\tau$ in that direction based on a fixed hyperparameter $\kappa$, which could in principle lead to under- or over-shooting. Our proposed approach, meanwhile, is guaranteed to move the activations to the closest point in $\mathcal R^t$. This highlights perhaps the greatest strength and weakness of the approach---it relies on the extent to which the language classifier used is able to identify causally-relevant language-specific regions. However, it also means that it is possible to use different classifiers and compare their performance, and given the theoretical guarantees, provides a method for directly testing whether the decision boundary learned by a given classifier is causally relevant.

A more extensive treatment of the \textsc{LiSeCo} method, including a brief application to the target-language generation task, can be found in \citet{cheng2026liseco}.

\section{Exploratory Analyses}
\label{sec:preliminary_analyses}
We use the evaluation framework proposed in \Cref{sec:evaluation} to evaluate the LiSeCo-based approach proposed in \Cref{sec:multilingual_control}. We compare this to two baselines---\textsc{DiffMeans} and \textsc{ActAdd}.

\subsection{Method}
\subsubsection{Task}
\textsc{StoryCloze} \citep{mostafazadeh_2016_CorpusClozeEvaluation} is a benchmark consisting of stories with two possible endings, one plausible and one implausible, which has been translated into multiple languages as \textsc{X\textsc{StoryCloze}} \citep{lin_few-shot_2022}. We construct a crosslingual variant of the task where the input is in the source language and the two possible endings are in the target language. The standard version of the task is to test whether a given language model assigns a higher probability to the plausible continuation. Because we study how activation steering shapes generation, we let the model generate the final sentence and evaluate it as described below.

\subsubsection{Activation Steering}
For \textsc{LiSeCo}, we use a linear classifier to compute the boundary between the language-specific region of each language based on the activations at the last word of paired sentences from the FLORES development set (997 items; \citealp{goyal_flores-101_2022}). For \textsc{DiffMeans}, we calculate the difference between the mean activations (at the last word) of the FLORES sentences in each language for each layer. \textsc{ActAdd} follows the same approach as \textsc{DiffMeans}, but the difference is calculated between two specific words; in this case, the difference between the activations of the endonyms of each language, e.g., `English' for English or `{\dn Eh\306wdF}' for Hindi. 

\textsc{LiSeCo} is formulated such that it can be applied at all layers \citep{cheng2024linearly}; but this is not true for \textsc{ActAdd} and \textsc{DiffMeans}---in both cases, the intervention is generally only applied at one or a few layers, determined empirically based on performance \citep{marks_geometry_2024,turner2023activation,lu_paths_2025}. We thus evaluate the performance of the \textsc{ActAdd} and \textsc{DiffMeans} methods for interventions at each layer $t$. In addition, because output language is generally thought to be controlled by the later layers (see \S\ref{sec:related_work}), we also try an alternative approach, where we apply the intervention at all layers from a given layer $t$ to the last layer. To allow full comparison we also apply \textsc{LiSeCo} under both of these settings. In addition to layer, each method also includes another tunable hyperparameter as discussed below.

\textsc{LiSeCo} guarantees a representation that lies within probability range $[\alpha^{\text{min}}, \alpha^{\text{max}}]$. For example, in a case where the classifier assigns English the 0 label and Spanish the 1 label, we can set the bounds $[0,0.1]$ to restrict the probability of the generation being English (as determined by the classifier) to $p\leq0.1$ and Spanish to $p>0.9$, and vice-versa by setting the bounds to $[0.9,1]$. Thus, we manipulate this `inner' bound $\alpha^{\text{inner}}$ (i.e., 0.1 or 0.9 in the aforementioned examples). With \textsc{ActAdd}, the difference between the representations is multiplied by an `injection coefficient' $c$ (\citealp{turner2023activation}; corresponding to $\kappa$ in \S\ref{sec:multilingual_control}) before being applied to the model. With \textsc{DiffMeans}, as with \textsc{ActAdd}, differences are multiplied by a value before being applied to the model. For consistency, we also refer to this as the injection coefficient $c$ in the case of \textsc{DiffMeans}.

\begin{figure}[h]
    \centering
    \includegraphics[width=\linewidth]{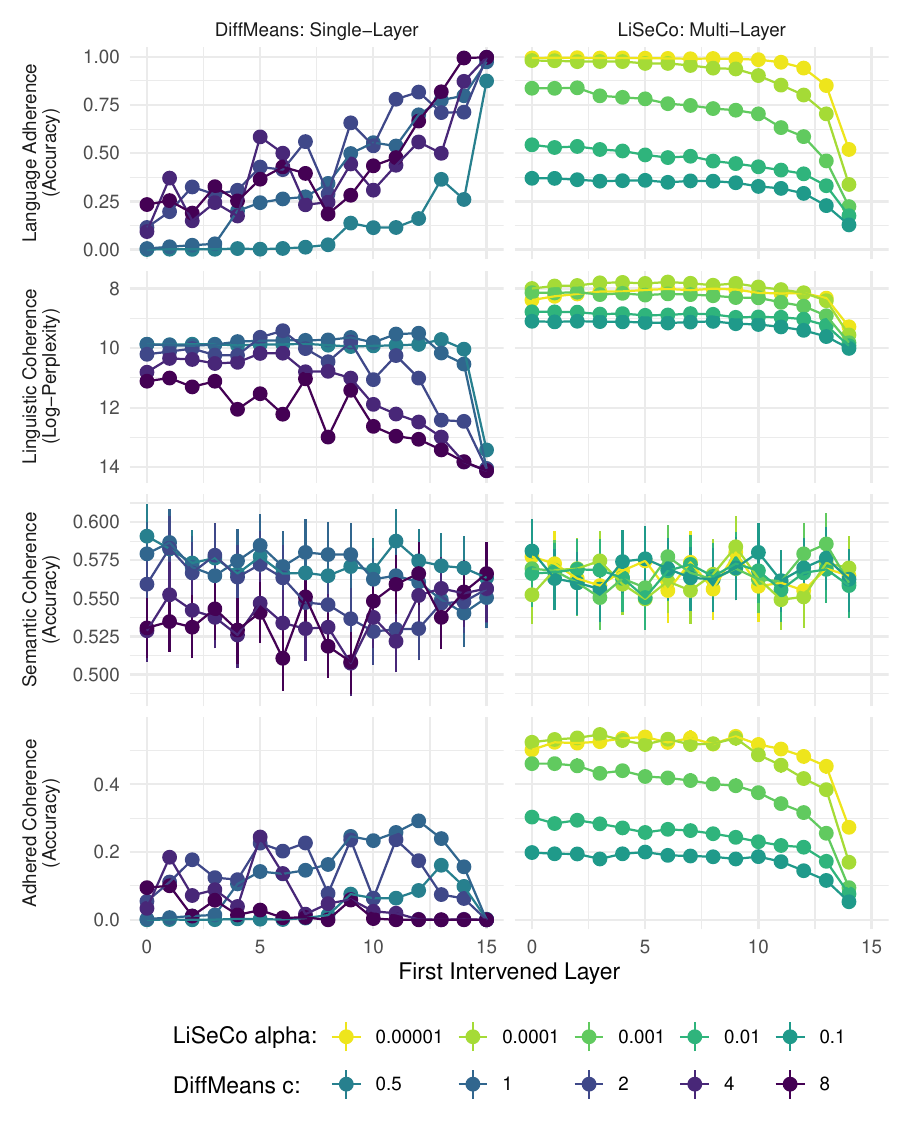}
    \caption{Results of the exploratory analyses on Llama 3.2 1B. Points represent mean adhered coherence based intervention starting layer, and error bars represent 95\% confidence intervals based on bootstrapping.}
    \label{fig:train_evals_summary}
\end{figure}

\subsubsection{Model and Settings}

Testing all possible settings of all steering methods is not computationally feasible, so we carry out exploratory experiments on Llama 3.2 1B \citep{grattafiori_llama_2024,meta_ai_llama_2024} to narrow down the scope of our analyses. We select the three \textsc{X\textsc{StoryCloze}} languages that are officially supported by the model: English (\texttt{eng}), Spanish (\texttt{spa}), and Hindi (\texttt{hin}). We evaluate on all 6 language pairs: English source with Spanish target (\texttt{eng}$\rightarrow$\texttt{spa}), \texttt{eng}$\rightarrow$\texttt{hin}, \texttt{hin}$\rightarrow$\texttt{eng}, \texttt{hin}$\rightarrow$\texttt{spa}, \texttt{spa}$\rightarrow$\texttt{eng}, and \texttt{spa}$\rightarrow$\texttt{hin}. For \textsc{LiSeCo}, we set $\alpha^{\text{inner}} \in \{10^{-1},10^{-2},10^{-3},10^{-4},10^{-5}\}$; for \textsc{ActAdd} and \textsc{DiffMeans}, $c \in \{0.5,1,2,4,8\}$.

\begin{figure*}[t!]
    \centering
    \includegraphics[width=\linewidth]{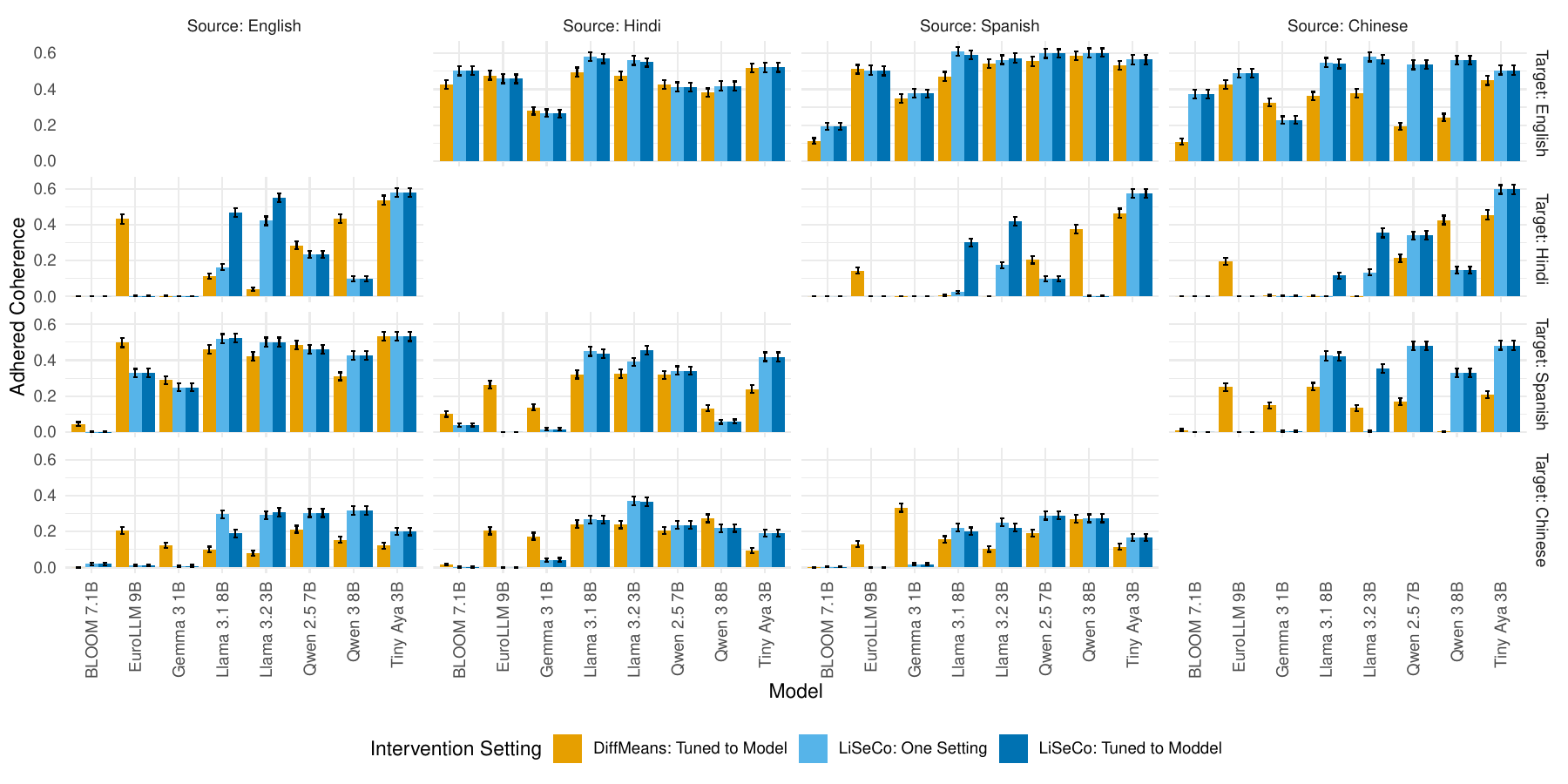}
    \caption{Test set results for all language pairs. Error bars reflect 95\% bootstrapped confidence intervals.}
    \label{fig:liseco_diffmeans_lang}
\end{figure*}

\subsubsection{Evaluation}
We evaluate the target-language generations based on the language adherence, linguistic coherence, and semantic coherence metrics defined in \Cref{sec:evaluation}, as well as the combined adhered coherence metric. We evaluate \textbf{language adherence} using the fastText language ID model released as part of the \textit{No Language Left Behind} project \citep{costa-jussa_no_2022}. We evaluate \textbf{semantic coherence} as the cosine similarity between the embedding of the generated sentence and each of the two reference \textsc{XStoryCloze} sentences using the Language-Agnostic BERT Sentence Embeddings model (LaBSE; \citealp{feng_2022_LanguageagnosticBERTSentence}), considering a generation closer to the plausible continuation than the implausible continuation to be correct. Finally, we evaluate \textbf{linguistic coherence} based on perplexity calculated using the $n$-gram language models trained by \citet{gonzalez_ponferrada_edugpkenlm_2024}. We calculate \textbf{adhered coherence} as discussed in \Cref{sec:evaluation}.

\subsection{Results}

We show a summary of our results in \Cref{fig:summary_fig}A. As can be seen, multi-layer \textsc{LiSeCo} and single-layer \textsc{DiffMeans} perform substantially better than the other activation steering methods on average (across all hyperparameters and language pairs); thus, we focus on these for the remainder of the paper. In \Cref{fig:liseco_diffmeans_lang}, we show the language adherence, linguistic coherence, and semantic coherence achieved using each of these methods, and how these are shaped by the first intervention layer and their method-specific hyperparameter (the full breakdown by language pair is provided in \Cref{app:prelim_full}). First, we observe that \textsc{LiSeCo} generally performs better than \textsc{DiffMeans}, especially for smaller values of $\alpha^{\text{inner}}$. We additionally observe that \textsc{DiffMeans} performs best when it is applied with $c=1$ to one of the final third of layers; \textsc{LiSeCo} performs better when applied with a very low $\alpha^{\text{inner}}$ to all or most layers.

\section{Targeted Analyses}
Given the results of \S\ref{sec:preliminary_analyses}, we carry out our main experiments on single-layer \textsc{DiffMeans} and multi-layer \textsc{LiSeCo} only. We follow the method in \S\ref{sec:preliminary_analyses} with a more restricted set of hyperparameters (see \S\ref{sssec:hyperparam}) on a larger set of (generally larger) models: BLOOM 7.1B, EuroLLM 9B, Gemma-3 1B, Llama-3.1 8B, Llama-3.2 3B, Qwen-2.5 7B, Qwen-3 8B, and Tiny-Aya-Base. After selecting the best hyperparameters for each method, we compare the performance of the two steering methods on the held-out test set of 1,512 items for each language pair. To test hyperparameter generalization, we add Chinese (\texttt{zho}) as an additional language\footnote{Note that Llama 3.1/3.2 models do not officially support Chinese, so we only consider this for research purposes.} and test all the pairs it forms with \{\texttt{eng}, \texttt{spa}, \texttt{hin}\}.

\subsection{Hyperparameter Selection}
\label{sssec:hyperparam}
We consider two different hyperparameter approaches for \textsc{DiffMeans} and \textsc{LiSeCo}. 

\textbf{\textsc{DiffMeans}} was shown in \S\ref{sec:preliminary_analyses} to work best when applied to one of the final third of layers with $c=1$. Thus, we fix $c=1$, and use the training set to select the best layer for each model.

\textbf{\textsc{LiSeCo}} was shown in \S\ref{sec:preliminary_analyses} to perform well when applied to all hidden layers with a low $\alpha^{\text{inner}}$. We thus apply \textsc{LiSeCo} to all layers in $[1,\max{(t)}]$ (i.e., all but the static embedding layer; as in \citealp{cheng2024linearly}) and vary $\alpha^{\text{inner}}$. Since our best-performing values were the lowest (i.e., $\alpha^{\text{inner}} \in \{10^{-4},10^{-5}\}$), we also include $\alpha^{\text{inner}} \in \{10^{-6},10^{-7},10^{-8}\}$. We tune $\alpha^{\text{inner}}$ to each model, and also find the best $\alpha^{\text{inner}}$ across all models.

\subsection{Results}
Our model-level results are presented in \Cref{fig:summary_fig} (full results including each metric for each language pair are provided in \Cref{app:full_experimental_results}). We observe that on the whole, our proposed method, \textsc{LiSeCo}, performs at least as well as \textsc{DiffMeans}: both \textsc{LiSeCo} settings out-perform model-tuned \textsc{DiffMeans} when applied to BLOOM, Llama 3.1, Llama 3.2, Qwen 2.5, and Aya. In the case of Qwen 3, we do not see a clear difference in either direction, and we see better \textsc{DiffMeans} performance for EuroLLM and Gemma 3. When we further break down the results according to language pair (\Cref{fig:liseco_diffmeans_lang}), we see that these patterns are non-uniform. For example, we see that for EuroLLM, \textsc{LiSeCo} only works for English targets (i.e., \texttt{x}$\rightarrow$\texttt{eng}) and for \texttt{eng}$\rightarrow$\texttt{spa}, and Gemma shows a similar but less extreme pattern. Another case of asymmetry is that \textsc{DiffMeans} works better than \textsc{LiSeCo} for Qwen 3 when the target language is Hindi (i.e., \texttt{x}$\rightarrow$\texttt{hin}), but the reverse is true for the Llama models. We also observe that a single \textsc{LiSeCo} setting ($\alpha^{\text{inner}}=10^{-8}$) works best for most models, with the exception of Llama 3.1 8B ($\alpha^{\text{inner}}=10^{-5}$) and Llama 3.2 3B ($\alpha^{\text{inner}}=10^{-6}$), suggesting that there are commonalities in this across models at the task level.
 
\section{General Discussion}
Our study has several main takeaways. First, we provide a theoretical framework that characterizes optimal language control, and based on this, propose a \textsc{LiSeCo}-based method for activation steering of language. Unlike existing methods that use fixed-magnitude interventions, this approach dynamically computes the minimal intervention needed based on the current activation state, providing both theoretical guarantees and computational efficiency. The presented method is, to the best of our knowledge, the only one that balances three competing requirements: ensuring the perturbed activation lies within the target language region, minimizing disruption to the original representation, and is topology-aware with embedding space geometry (i.e. it only moves activations along the axis orthogonal to the decision boundary between language-specific regions).

Second, we provide what is to our knowledge some of the first quantitative evidence that activation steering alone can be used to control generation language while preserving linguistic and semantic coherence (for contemporaneous work, see \citealp{gurgurov-etal-2026-clas}). We demonstrate that the \textsc{LiSeCo}-based approach generally performs better than \textsc{DiffMeans} for 5 of the 8 models tested. Crucially, we observe that the performance of \textsc{LiSeCo} is relatively predictable and robust to hyperparameters---a single \textsc{LiSeCo} setting, namely, intervening on all but the static embedding layer with $\alpha^{\text{inner}}=10^{-8}$, performs better overall for 5 of the 8 models than \textsc{DiffMeans}, even when the latter method involves tuning the layer of the intervention to each model and the former does not (though this does also improve performance). The performance of \textsc{DiffMeans} and \textsc{ActAdd} is far more sensitive to their hyperparameters, especially at the language-pair level (see Appendix \ref{app:prelim_full}).

Finally, we consider the cases where \textsc{LiSeCo} under-performs relative to \textsc{DiffMeans}. Crucially, the asymmetries in performance (e.g., \texttt{eng}$\rightarrow$\texttt{hin} vs. \texttt{hin}$\rightarrow$\texttt{eng}) suggest that \textsc{LiSeCo} performance may arise from poor classifier performance. For example, language subspaces identified by the classifier may not be causally implicated; that is, it is possible to train an accurate linear classifier probe to separate the representations of two languages, but this does not mean that manipulating the language model state to fall within these regions will cause the language model to generate in the corresponding language. This highlights the importance of causally testing probing results \citep[see][]{elazar_2021_AmnesicProbingBehavioral,ravichander_2021_ProbingProbingParadigm,kumar_2022_ProbingClassifiersAre}.

We also observe strong asymmetries. For example, all models are able to be steered into generating English with \textsc{LiSeCo}, but EuroLLM and Gemma struggle to be steered to other languages \textit{from} English. One possible explanation for this is an inherent feature of any binary classification approach in a context where there are more than two classes (in this case, languages in multilingual models): one or both of the regions on either side of the decision boundary will include model activation states that correspond to other languages that the models are trained on (i.e., those not being classified). Given that English is almost always the primary training language, we might expect most language models are likely to have higher-quality representations of English than its paired languages, and English may occupy ``more'' of the total representation space \citep[see, e.g.,][]{wendler2024llamas,schut_2025_MultilingualLLMsThink,zhong_2025_WhatLanguageNonEnglishCentric,lu_paths_2025}; this may make it easier to delineate English from non-English model states. Crucially, however, \textsc{LiSeCo} is compatible with a wide range of possible linear classifiers; and thus, there is substantial scope for improvement. Furthermore, given the theoretical guarantees of the method, such improvements are likely to aid in uncovering the nature of the representations in language models that causally shape generation language.

\section*{Limitations}
Our main limitation in this work is coverage. Due to the computational cost of the experiments ($\sim$7000 GPU hours on a cluster including L40S, A100, and H100 GPUs), we chose to evaluate 8 language models on one task made up of 12 language pairs.

We do not believe that the number of models is a substantial concern; in fact, most previous work in this area involves a smaller number of models (see, e.g., \citealp{chang_2022_GeometryMultilingualLanguage,gurgurov2025language,sundar2025steering,mahmoud2025improving,lu_paths_2025,wang2025bridging,sterz_2025_ReCoVeRTargetLanguage,lopo_2025_LanguageSurgeryMultilingual}). Nonetheless, a larger number of models would allow for more robust generalizations to be made.

Similarly, while we believe that 12 language pairs (including three scripts and two language families) allows us to answer our research questions sufficiently, a larger and more diverse sample of languages would enable more robust conclusions, and would allow potentially informative comparisons between pairs of languages that are related or typologically similar to different degrees.

Finally, we construct a specific variant of the \textsc{X\textsc{StoryCloze}} task in order to avoid potential confounds arising from prompt adherence; this allows us to directly test the efficacy of different steering approaches. However, a more complete picture would be provided by considering a wider range of tasks, including those where the prompt explicitly states the target language (e.g., translation).

\section*{Ethical Considerations}
We do not consider our work to present any substantial risks. However, since activation steering methods edit the activations of models directly, it is possible that model generations may be more unstable and unpredictable, which in principle could increase the risk of harmful content. Thus, without further safety measures, the intended use for our work is limited to fundamental research.

With respect to licenses and intended use, we report the details of the main scientific artifacts used below:

\paragraph{Datasets:}
\begin{itemize}
    \item FLORES \citep{goyal_flores-101_2022}: CC BY-SA 4.0
    \item XStoryCloze \citep{lin_few-shot_2022}: CC BY-SA 4.0
\end{itemize}

\paragraph{Models:}
\begin{itemize}
    \item fastText LangID \citep{costa-jussa_scaling_2024}: CC-BY-NC-4.0 License
    \item BLOOM 7.1B \citep{bigscienceworkshop_2023_BLOOM176BParameterOpenAccess}: BigScience RAIL License v1.0
    \item EuroLLM 9B \citep{martins_2025_EuroLLMMultilingualLanguage}: Apache 2.0 License
    \item Gemma-3 1B Pretrained \citep{team_2025_Gemma3Technical}: Gemma License
    \item Llama-3.1 8B \citep{grattafiori_llama_2024}: Llama 3.1 License
    \item Llama-3.2 3B \citep{grattafiori_llama_2024}: Llama 3.2 License
    \item Qwen-2.5 7B \citep{qwenteam_2025_Qwen25TechnicalReport}: Apache 2.0 License
    \item Qwen-3 8B Base \citep{yang_2025_Qwen3TechnicalReport}: Apache 2.0 License
    \item Tiny-Aya-Base \citep{salamanca_2026_TinyAyaBridging}: CC-BY-NC-4.0 License
    \item $n$-gram models \citep{gonzalez_ponferrada_edugpkenlm_2024}: MIT License
    \item LaBSE \citep{feng_2022_LanguageagnosticBERTSentence}: Apache 2.0 License
\end{itemize}

\paragraph{Research code:}
\begin{itemize}
    \item LiSeCo and ActAdd implementations from \citet{cheng2024linearly}: \url{https://github.com/chengemily1/llm-control}; no license stated; provided for research purposes.
\end{itemize}

\section*{Acknowledgments}
James Michaelov was supported by a grant from the Andrew W. Mellon foundation (\#2210-13947) during the writing of this paper.

\bibliography{custom}

\clearpage
\onecolumn
\appendix

\section{Exploratory Analyses: Further Details}
\label{app:prelim_full}

\begin{figure}[H]
    \centering
    \includegraphics[width=\linewidth]{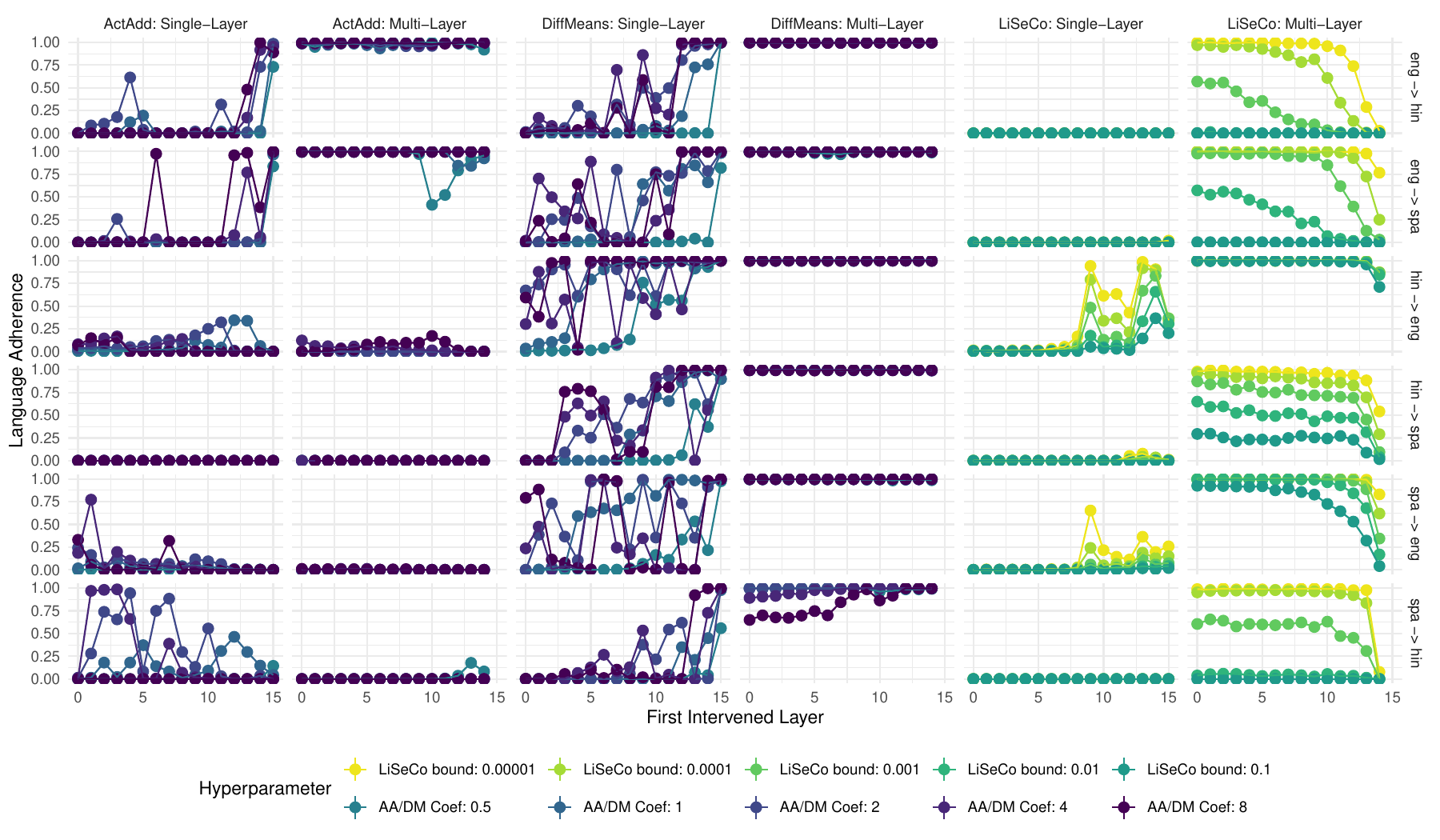}
    \caption{Llama 3.1 1B language adherence for all evaluated activation steering configurations and language pairs on the \textsc{XStoryCloze} training set.}
    \label{fig:train_evals_full_a}
\end{figure}

\begin{figure}[H]
    \centering
    \includegraphics[width=\linewidth]{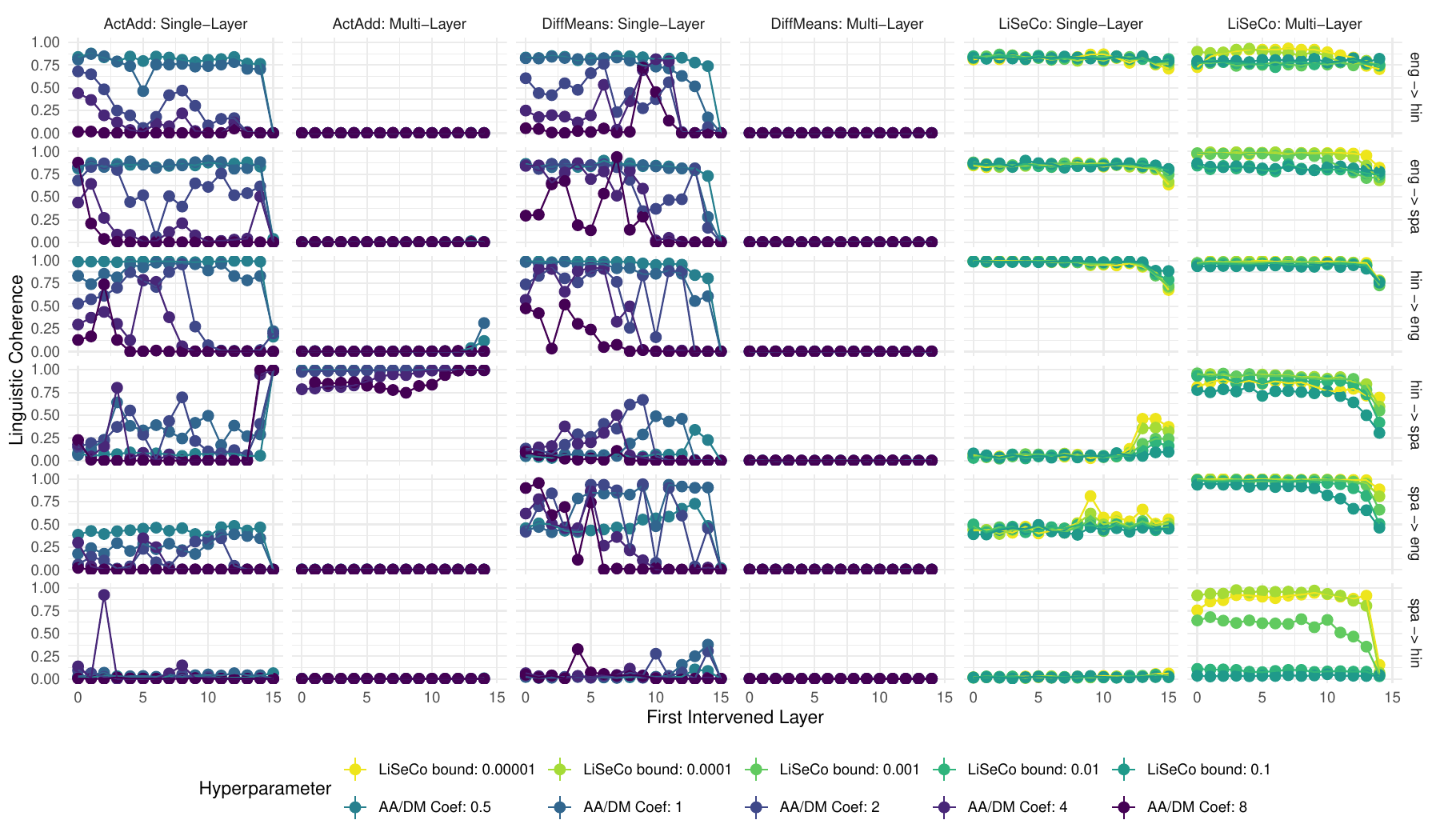}
  \caption{Llama 3.1 1B linguistic coherence for all evaluated activation steering configurations and language pairs on the \textsc{XStoryCloze} training set.}
    \label{fig:train_evals_full_l}
\end{figure}

\begin{figure}[H]
    \centering
    \includegraphics[width=\linewidth]{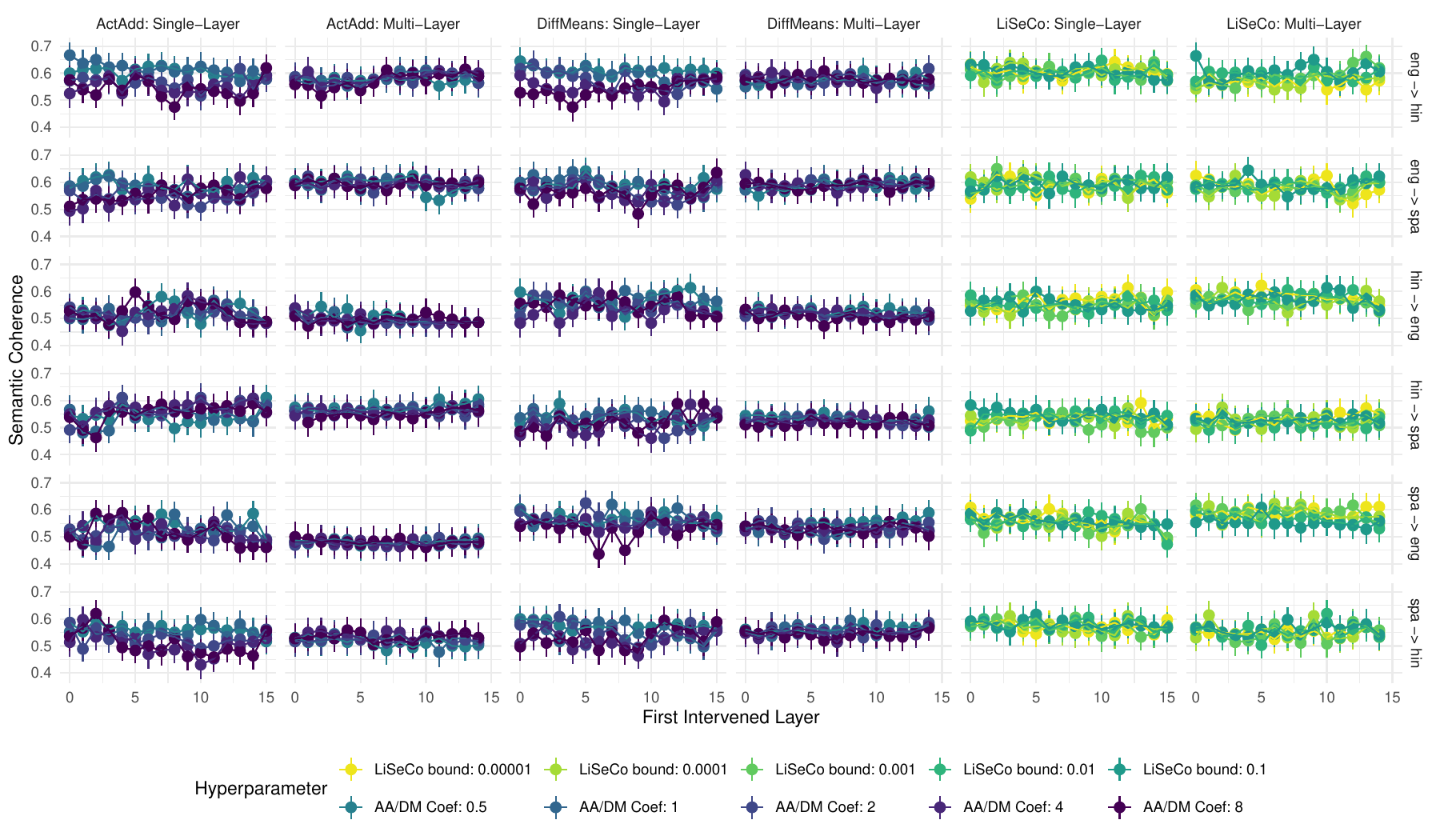}
  \caption{Llama 3.1 1B semantic coherence for all evaluated activation steering configurations and language pairs on the \textsc{XStoryCloze} training set.}
    \label{fig:train_evals_full_s}
\end{figure}

\begin{figure}[H]
    \centering
    \includegraphics[width=\linewidth]{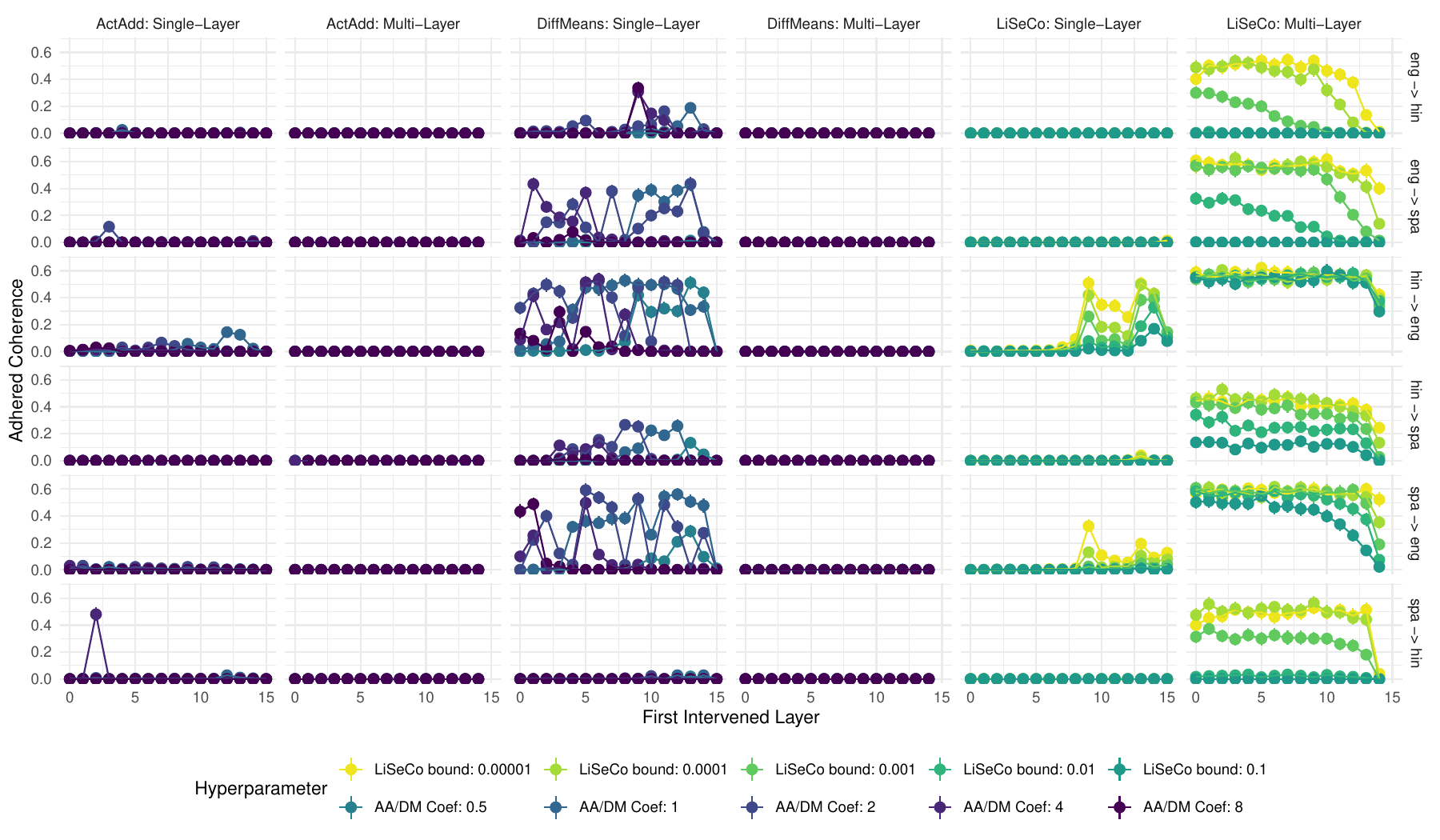}
  \caption{Llama 3.1 1B adhered coherence for all evaluated activation steering configurations and language pairs on the \textsc{XStoryCloze} training set.}
    \label{fig:train_evals_full_ac}
\end{figure}

We provide the full details of Llama 3.1 1B language adherence (\Cref{fig:train_evals_full_a}), linguistic coherence (\Cref{fig:train_evals_full_l}), semantic coherence (\Cref{fig:train_evals_full_s}), and adhered coherence (\Cref{fig:train_evals_full_ac}) broken down by steering method, set of intervened layers, method hyperparameter, and language pair.

\clearpage
\section{Targeted Analyses: Further Details}
\label{app:full_experimental_results}
\begin{figure}[H]
    \centering
    \includegraphics[width=\linewidth]{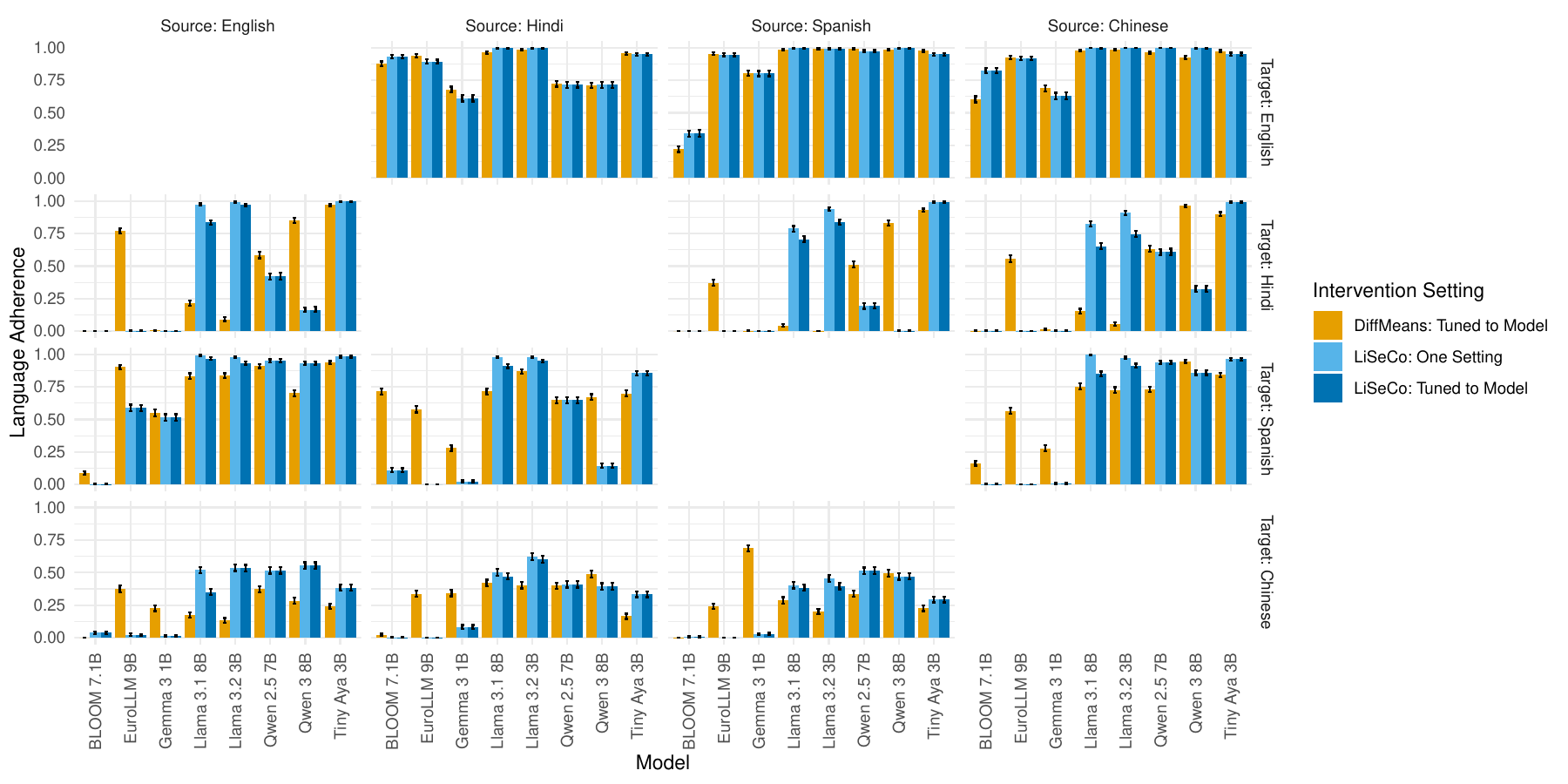}
    \caption{Language Adherence of all language models and language pairs on the \textsc{XStoryCloze} test set.}
    \label{fig:test_evals_full_a}
\end{figure}

\begin{figure}[H]
    \centering
    \includegraphics[width=\linewidth]{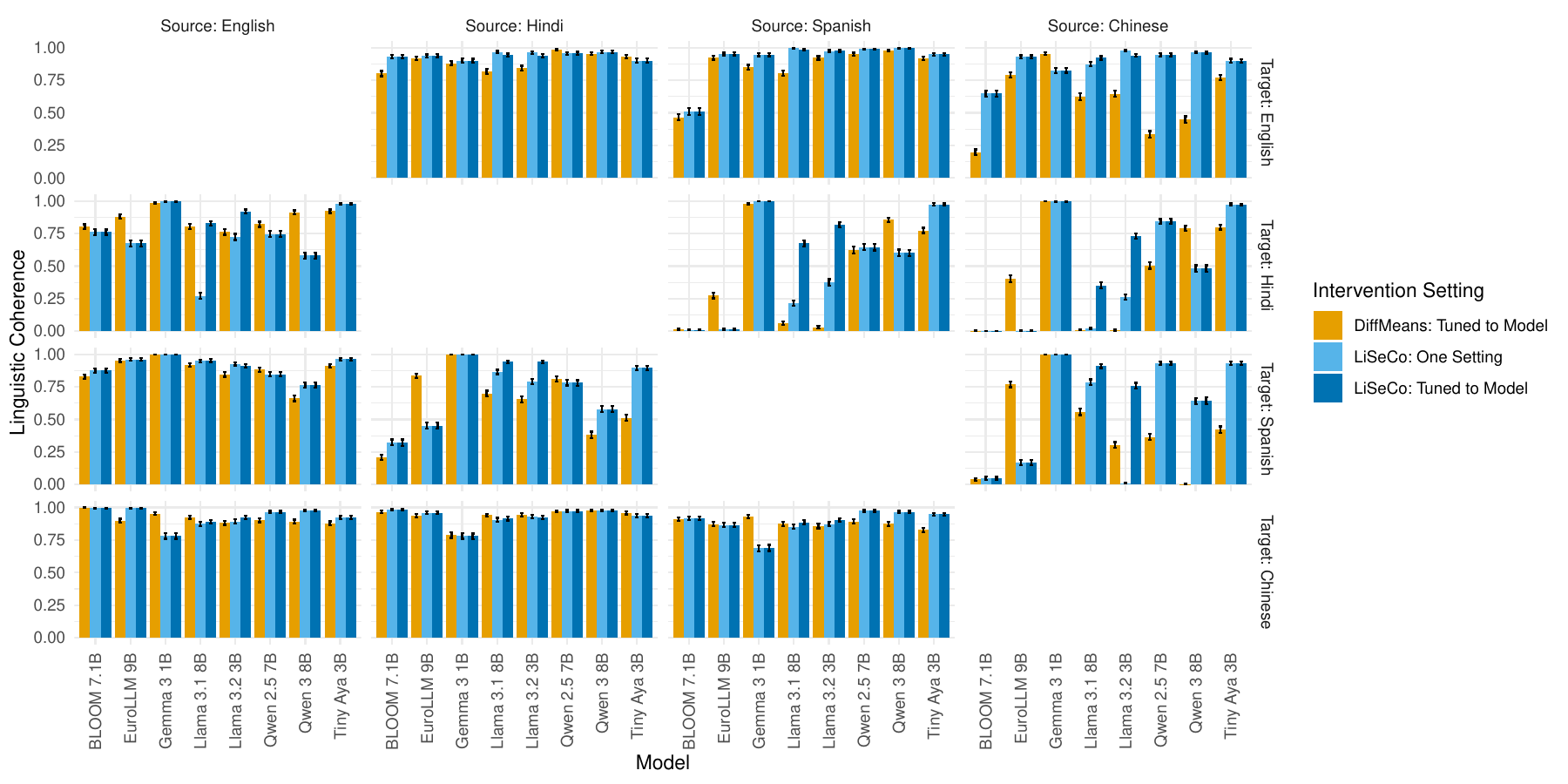}
  \caption{Linguistic Coherence of all language models and language pairs on the \textsc{XStoryCloze} test set.}
    \label{fig:test_evals_full_l}
\end{figure}

\begin{figure}[H]
    \centering
    \includegraphics[width=\linewidth]{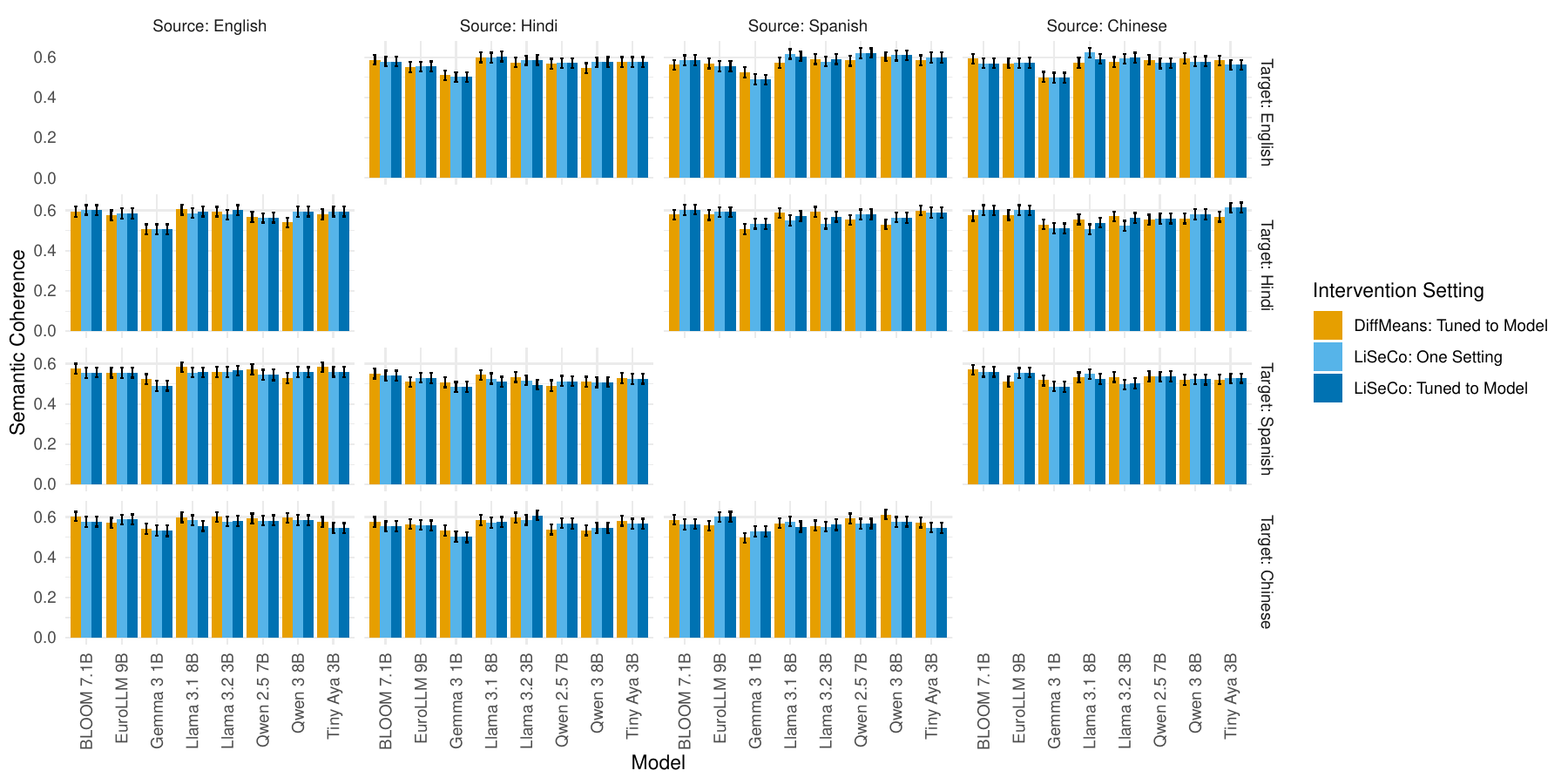}
  \caption{Semantic coherence of all language models and language pairs on the \textsc{XStoryCloze} test set.}
    \label{fig:test_evals_full_s}
\end{figure}

\begin{figure}[H]
    \centering
    \includegraphics[width=\linewidth]{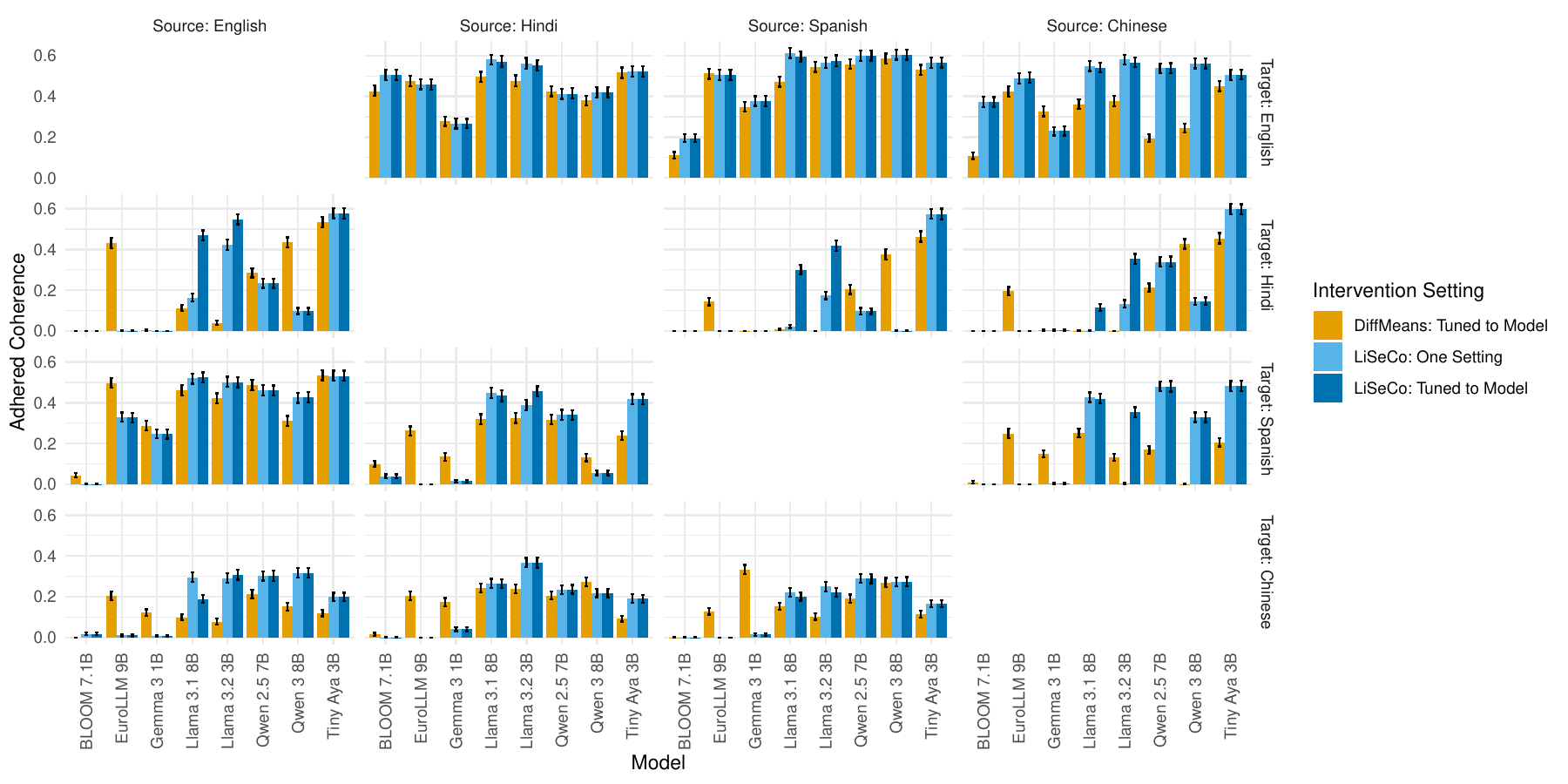}
  \caption{Adhered Coherence of all language models and language pairs on the \textsc{XStoryCloze} test set.}
    \label{fig:test_evals_full_ac}
\end{figure}

We provide the test set language adherence (\Cref{fig:test_evals_full_a}), linguistic coherence (\Cref{fig:test_evals_full_l}), semantic coherence (\Cref{fig:test_evals_full_s}), and adhered coherence  (\Cref{fig:test_evals_full_ac}) for each language model on each language pair below.

\end{document}